\def\year{2018}\relax
\documentclass[letterpaper]{article} %DO NOT CHANGE THIS
\usepackage{aaai18}  %Required
\usepackage{times}  %Required
\usepackage{helvet}  %Required
\usepackage{courier}  %Required
\usepackage{url}  %Required
\usepackage{graphicx}  %Required
\usepackage{mathptmx}
\usepackage{graphicx}
\usepackage{amsmath}
\usepackage{amsfonts}
\usepackage{booktabs}
\usepackage{natbib}
\begin{document}
\title{Encoding Temporal Markov Dynamics in Graph  \\ for Visualizing and Mining Time Series}
\author{Lu Liu \\ 
Department of Computer Science and Electric Engineering \\
University of Maryland Baltimore County \\
liulumy813@gmail.com \\
\And Zhiguang Wang \\  
Microsoft Corporation \\
zhiguang.wang@microsoft.com
}

\maketitle
\begin{abstract}
	Time series and signals are attracting more attention across statistics, machine learning and pattern recognition as it appears widely in the industry， especially in sensor and IoT related research and applications, but few advances has been achieved in effective time series visual analytics and interaction due to its temporal dimensionality and complex dynamics. Inspired by recent effort on using network metrics to characterize time series for classification, we present an approach to visualize time series as complex networks based on the first order Markov process in its temporal ordering. In contrast to the classical bar charts, line plots and other statistics based graph, our approach delivers more intuitive visualization that better preserves both the temporal dependency and frequency structures. It provides a natural inverse operation to map the graph back to raw signals, making it possible to use graph statistics to characterize time series for better visual exploration and statistical analysis. Our experimental results suggest the effectiveness on various tasks such as pattern discovery and classification on both synthetic and the real time series and sensor data.
\end{abstract}
\section{Introduction}

\maketitle
Time series data is ubiquitous in both industry and academia. Except for the perceptual signals that we are able to easily understand such as audio, video and nature language, most of the other signals are generated from sensors. Therefore, Understanding temporal patterns is key to gaining knowledge and insight from those sensor data.

Collectively, our ability to store data from sensors now far exceeds
the rate at which we are able to understand it. Scientists and domain experts expect to discover patterns, thus to help the further understanding of those mechanical signals in various tasks like classification and anomaly detection. A lot of advances have been made to learn the temporal correlations that are often embedded in time series. Visualization of time series provides a new perspective to explore those complex dynamics in a more intuitive way and helps to improve the interaction between signals and users. 

Many visualization methods have been introduced to address time series mining tasks. The line graph is widely used \cite{javed2010graphical,mclachlan2008liverac}. While the standard time series graph is effective when dealing with a small data space, it is
more challenging to perform common tasks on larger data. Pixel plot is explored to represent a time-series as an arrangement of pixels encoded with different hues which encode the underlying data \cite{Kincaid:2006:LGE:1133265.1133348,Hao:2007:MTV:2384179.2384183}. Besides aggregating the data into temporal segments with lens view, layout based techniques provide a linear mapping by modifying the spatial arrangement of the time while transforming those graphs which enhance the display.% In \cite{hao2005importance} the authors propose an early importance driven layout scheme while \cite{javed2010stack} presents a multi-focus zooming approach which maintains context and temporal distance zooming. In contrast to these schemes, \cite{javed2013stack} compares stack zooming against standard techniques for navigation of temporal data.

More complex and noisy real-world time series data is often difficult to observe through visualization because the dynamics are either too complex or unintuitive. To provide more profound insights about the temporal dynamics embedded in time series, one possible solution is to reformulating the data to explicitly extract features for better visual encoding of the temporal dependency.

Reformulating the features of time series as visual clues has been studied in physics as a kind of dynamical systems and raised much attention in data mining and machine learning research. One major approach is to build different network structures from time series for visual analytics. Recurrence Networks were proposed to analyze the structural properties of time series from complex systems \cite{donner2010recurrence}. They build adjacency matrices from the predefined recurrence functions to interpret the time series as complex networks for visual inspection and metric based classification. \cite{silva2013time} extended the recurrence plot paradigm for time series classification using compression distance. Another way to build a weighted adjacency matrix is to model the temporal-spatial transition dynamics with first order Markov matrix \cite{campanharo2011duality}. Although these maps demonstrate distinct topological properties among different time series to visualize the intrinsic statistics, it remains unclear how these topological properties relate to the original time series to help visual analytics as they have no exact inverse operations. 

\cite{wang2015imaging} proposed an encoding framework to transform time series to images as the input for deep convolution neural networks and achieved state-of-the-art classification performance. The proposed temporal Markov encoding maps have exact/approximate inverse maps, making these transformations more interpretable with graph visualization techniques. 

Inspired by the temporal Markov encoding approaches, we design a novel visualization framework to translate time series into complex networks which preserves both temporal ordering and statistical dynamics. Our encoding map also has a natural inverse, which makes it possible to visually explore the interesting patterns in the generated graph and match them back to the original data. Our experimental results suggest the effectiveness on various tasks such as pattern discovery, classification and anomaly detection on both benchmark and real time series data.

\section{Data Transformation  and Quantitative Encoding}
\begin{figure*}[t]
	\centering
	\includegraphics[width=0.8\textwidth]{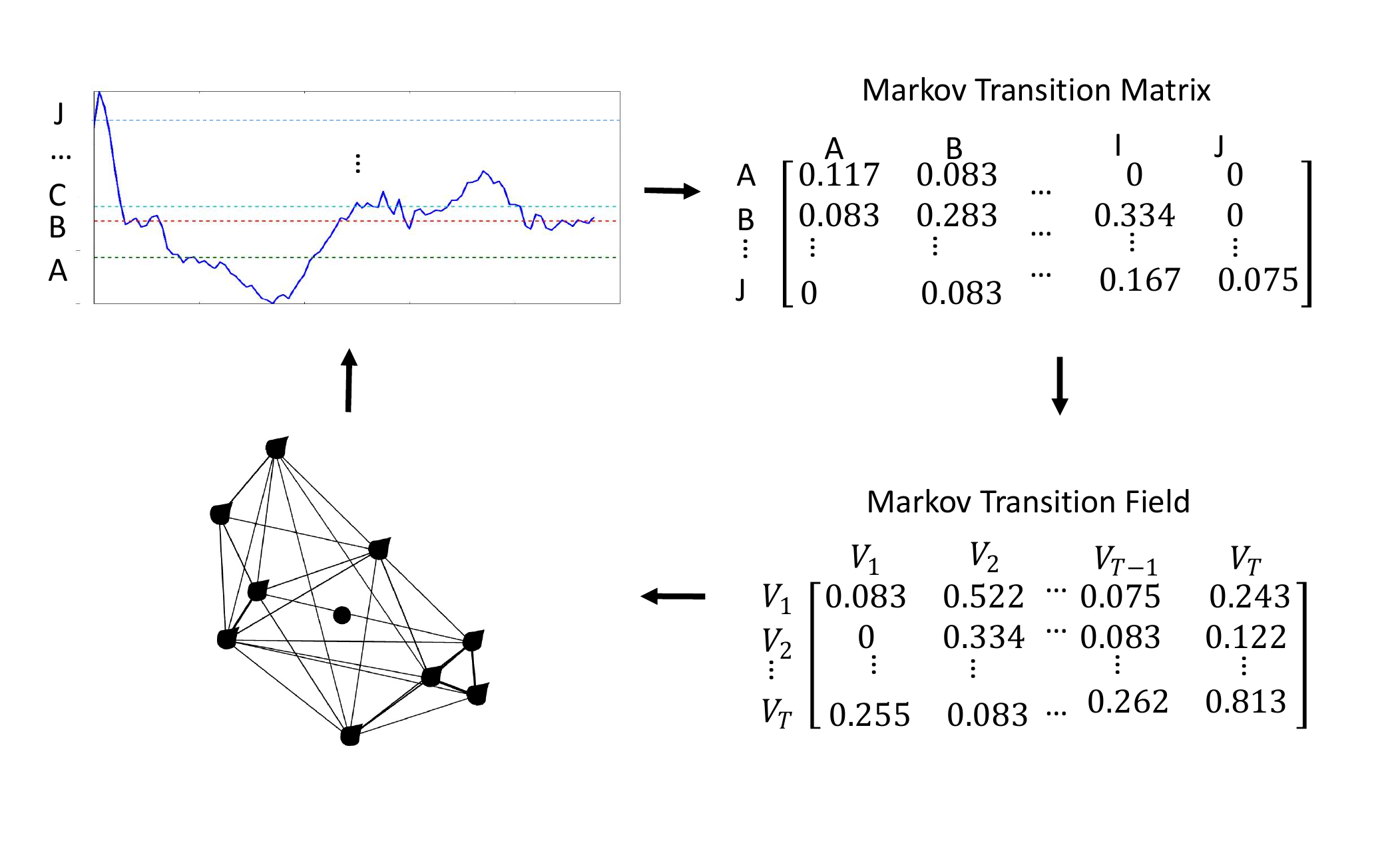}
	\caption{Illustration of the proposed encoding map of a Markov Transition Field. $X$ is a sequence of the  typical time series in the 'ECG' dataset. $X$ is first discretized into $Q$ quantile bins. Then we calculate its Markov Transition Matrix $W$ and finally build its MTF $M$ by Equation. (\ref{eqn:MTF}). We reduce the image size from $96 \times 96$ to $48 \times 48$ by averaging the values in each non-overlapping $2 \times 2$ patch.}
	\label{fig:encodingTS2MTF}
\end{figure*}

Temporal and frequency correlation are major dependencies embedded in time series data. To build a comprehensive but intuitive visualization, the extracted features from the designed data transformation framework should be able to represent the dynamics in both time and frequencies while there exists a reverse operation to map the information back to raw time series. In this section, we will introduce how to encode the dynamical frequency information in the temporal ordering step by step (Figure \ref{fig:encodingTS2MTF}). 

\subsection{Quantization and Markov Matrix}

This framework is similar to \cite{campanharo2011duality} for
encoding dynamical transition statistics. We develop that idea by
representing the Markov transition probabilities sequentially  to
preserve information in the temporal dimension. The first step is to quantize time series to build the first order Markov Matrix.

Given a time series $X = \{x_1, x_2, \cdots, x_n\}$, we need to quantize its values in $Q$ bins. Quantile is the most common way to achieve a discretized dictionary. By identifying the $Q$ quantile bins, each value $x_i$ is mapped to each $q_i$. In \cite{lin2003symbolic, lin2007experiencing}, a different quantization approach is proposed. Based on the empirical observation that normalized time series has a Gaussian distribution \cite{larsen1986introduction}, it is desirable to have a discretization technique that will produce bins with approximate equiprobability. We can simply determine the “breakpoints” that will produce an equal-sized areas under Gaussian curve. Breakpoints are a sorted list of numbers $Q = q_1, q_2, ..., q_Q$ such that the area under a $\mathnormal{N}(0,1)$ Gaussian curve follows $P(q_{i+1}) - P(q_i) = \frac{1}{Q}$.

Quantile mapping ignored the Gaussian priori, so it is more data-specific as the distribution on different datasets could drift significantly, although some data comes from the similar distribution. Gaussian mapping does not have such problems and performs well with Symbolic  Aggregate Approximation (SAX) in the classification and query tasks. In the next sections, we use Gaussian mapping to create the dictionary for quantizing time series. 

After assigning each $x_i$ to its corresponding bin $q_j$ ($j \in [1,Q]$), we construct a $Q \times Q$ weighted adjacency matrix $W$ by counting transitions among bins in the manner of a first-order Markov chain along each time step. $w_{i,j}$ is the frequency of a point in the bin $q_j$ followed by a point in the bin $q_i$. After normalization by $\sum_j{w_{ij}=1}$, $W$ is a Markov matrix:

\begin{equation}
	W = 
	\begin{bmatrix}
		w_{11|P(x_t \in q_1|x_{t-1} \in q_1)}  & \cdots & w_{1Q|P(x_t \in q_1|x_{t-1} \in q_Q)} \\
		w_{21|P(x_t \in q_2|x_{t-1} \in q_1)}  & \cdots & w_{2Q|P(x_t \in q_2|x_{t-1} \in q_Q)} \\
		\vdots  & \ddots & \vdots \\
		w_{Q1|P(x_t \in q_Q|x_{t-1} \in q_1)}  & \cdots & w_{QQ|P(x_t \in q_Q|x_{t-1} \in q_Q)} \\
	\end{bmatrix} 
	\label{eqn:MTP}
\end{equation}  

\subsection{Markov Transition Field and Network Graph}
Markov matrix, while incorporating the Markov dynamics, discards the conditional relationship between the distribution of $X$ and the temporal dependency on the time steps $t_i$.  Markov Transition Field (MTF) extends Markov matrix by aligning each probability along the temporal order. 

\begin{eqnarray}
	M =&
	\begin{bmatrix}
		M_{11}  & M_{12} & \cdots & M_{1n} \\
		M_{21}  & M_{22} & \cdots & M_{2n} \\
		\vdots  & \ddots & \vdots \\
		M_{n1}  & M_{n2} & \cdots & M_{nn} \\
	\end{bmatrix} \nonumber \\
	=&
	\begin{bmatrix}
		w_{ij|x_1 \in q_i,x_1 \in q_j}  & \cdots & w_{ij|x_1 \in q_i,x_n \in q_j} \\
		w_{ij|x_2 \in q_i,x_1 \in q_j}  & \cdots & w_{ij|x_2 \in q_i,x_n \in q_j} \\
		\vdots  & \ddots & \vdots \\
		w_{ij|x_n \in q_i,x_1 \in q_j}  & \cdots & w_{ij|x_n \in q_i,x_n \in q_j} \\
	\end{bmatrix} 
	\label{eqn:MTF}
\end{eqnarray} 

That is, in the Markov matrix $M$, the quantile bins that contain the data at time steps $i$ and $j$ (temporal axis) are $q_{i}$ and $q_{j}$ ($q \in [1,Q]$). $M_{ij}$ denotes the
transition probability of $q_{i} \rightarrow q_{j}$. In other words, we spread out matrix $W$, which contains the transition probability on the magnitude axis, into the $W$ by considering temporal positions.

By assigning the probability from the quantile at time step $i$ to the
quantile at time step $j$ at each value $M_{ij}$, the MTF $M$ encodes multi-scale transition probabilities of the time
series. $M_{i,j||i-j|=k}$ denotes the transition probability between
the points with time interval $k$. For example, $M_{ij|j-i=1}$
illustrates the transition process along the time axis with a skip
step. The main diagonal $M_{ii}$, which is a special case when $k=0$
captures the probability from each quantile to itself (the
self-transition probability). 

The square matrix $M$ can be interpreted as a network graph $G=(V, E)$ directly. Each value $M_{ij}$ represents the edge weights, each row/column corresponds to the vertex index. Note that   
such settings allow the inverse operation to match nodes back to the raw signals, as the mapping between node $V_i$ in the graph and the point $x_i$ in the signal is bijective. As in Figure \ref{fig:encodingTS2MTF}, the nodes in red, blue and yellow exactly match different time points respectively.

\begin{figure*}[t]
	\centering
	\includegraphics[width=0.85\textwidth]{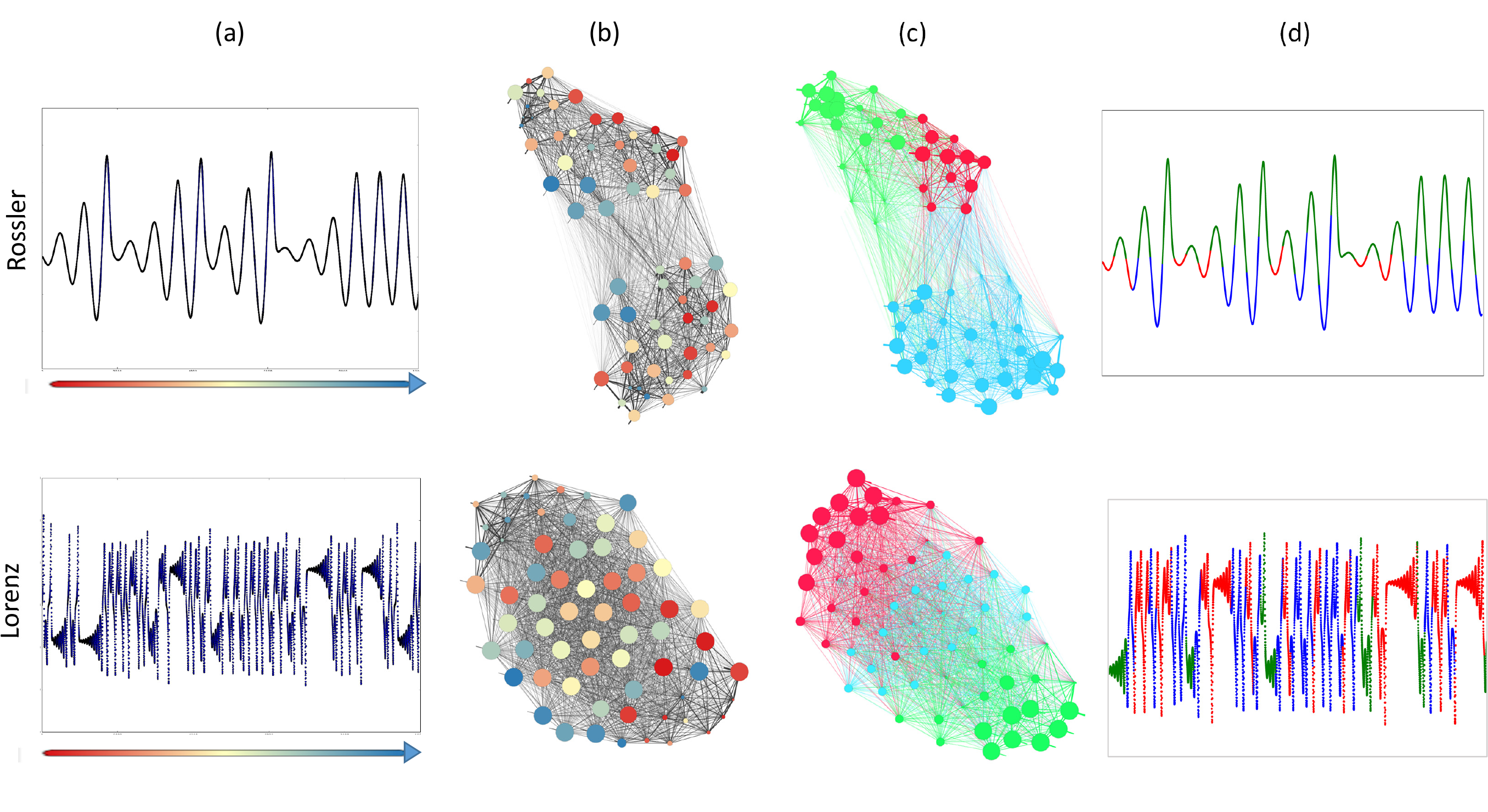}
	\caption{Network graph generated from Lorenz and Rossler system with $Q = 50$ and $S = 64$. (a). Original time series; (b). Graph with the flow encoding approach; (c). Graph with the modularity encoding approach (community resolution = 1); (d). Time series with the inverse mapping of the graph community.}
	\label{fig:Lorenz_Rossler}
\end{figure*}

\subsection{Dimension Reduction}
To make the matrix size manageable for efficient computation and visualization, two common dimension reduction techniques can be performed before or after the MTF encoding: 
\begin{itemize}
	\item \textbf{Piecewise Aggregate Approximation (PAA)} reduces the time series from $n$  dimensions to $w$ dimensions, the data is divided into $w$ equal sized “frames”. The mean values of the data falling within a frame is calculated as the compressed representation. The representation can be visualized as an attempt to approximate the original time series with a linear combination of linear basis functions. PAA is a special case of moving average model when the skip size equals to the length of the sliding window.
	\item \textbf{Blurring} on MTF is fulfilled by averaging the values in each non-overlapping $m \times m$ patch with
	a blurring kernel, typically a Bivariate Gaussian $\{\frac{1}{2\pi\sigma^2}e^{-\frac{x_1^2 + x_2^2}{2\sigma^2}}\}_{m \times m}$ or a simple average kernel $\{\frac{1}{m^2}\}_{m \times m}$. That is, we aggregate the transition probabilities in each subsequence of length $m$ together using a convolution operation with a smaller kernel. The final output size $S = \lceil \frac{n}{m} \rceil$.
\end{itemize}

PAA is simple and effective especially when the detailed structure of the time series can be ignored while the major trends/patterns are important, like in the tasks of classification and query. For analyzing the visual patterns, small information loss and minimum manipulation on the raw signal is preferred to preserve the information in the raw data. Hence, we only use Gaussian blurring to reduce the MTF size.

\section{Visual Encoding and Inverse Mapping}
The graph $G=(V,E)$ generated from MTF has a direct mapping from the vertex $V$ to the time index $i$ in the original time series. It is easy to encode the time index $\{1,2,...,n\}$ as the color of the vertices. Along the temporal axis, we choose PageRank \cite{page1999pagerank} weights as the vertex size and Markov matrix weight as the edge color to observe when (to which vertex) do the big information/citation flow occurs. 

Modularity is an important pattern in network analysis to identify the specific local structures. To verify the modules and distinct patterns, we also figure out the modularity label and the clustering coefficient \cite{blondel2008fast,latapy2008main} for each vertex and map them back to the original time series, and assign the modularity labels to each vertex.  In this way, another visual encoding framework is to encode the vertex/edge color as the module labels it belongs to and the vertex size as the clustering coefficient. So we have two visual encoding frameworks for general/flow inspection and modularity discovery (Table \ref{tab:visualencoding}).

\begin{table}[h]
	\centering
	\caption{Visual Encoding Framework}
	\begin{tabular}{rrr}
		\toprule
		\textit{Flow} & \textbf{Vertex} & \textbf{Edge} \\
		\textit{Encoding} & & \\
		\textbf{Color} & Time index & Markov matrix weight \\
		\textbf{Size}  & PageRank weight & Constant \\
		\midrule
		\textit{Modularity}  & \textbf{Vertex} & \textbf{Edge} \\
		\textit{Encoding} & & \\
		\textbf{Color} & Module label & Target module label  \\
		\textbf{Size}  & Clustering coefficient & Constant \\
		\bottomrule
	\end{tabular}%
	\label{tab:visualencoding}%
\end{table}%

As each vertex index and the time index are one-to-one mapping, we can easily map the interesting patterns we find in the graph back to the time series. As in Figure \ref{fig:Lorenz_Rossler} (c) and (d), the three communities discovered in both graph have their corresponding shapelets in the raw signal \cite{ye2009time}.  

\section{Experiments and Analysis}
\subsection{Pattern discovery}
%We apply the encoding approaches on two dynamical systems (Figure \ref{fig:Lorenz_Rossler} (a)). The first time series is generated by chaotic Lorenz equations
%
%\begin{eqnarray}
%	\frac{dx}{dt} &=& \sigma(-x+y) \nonumber \\
%	\frac{dy}{dt} &=& rx-y-xz \nonumber \\
%	\frac{dz}{dt} &=& bz+xy
%\end{eqnarray}
%
%with parameter values $\sigma=10, b = \frac{8}{3}$ and $r=28$. Numerical solutions of these equations leads to an attractor embedded in a three-dimensional space with coordinates. The trajectory rotates about one of two unstable fixed points then escapes to orbit the other fixed point. This behavior is recognizable in the variable $x$ since its values oscillate between the positive and the negative.
%
%The second time series is generated by the variable of the chaotic Rossler equations
%\begin{eqnarray}
%	\frac{dx}{dt} &=& -(y+z) \nonumber \\
%	\frac{dy}{dt} &=& x+ay \nonumber \\
%	\frac{dz}{dt} &=& b+z(x-c)
%\end{eqnarray}
%
%with parameter values $a=0.432$, $b = 2$ and $c=4$. The trajectory within the attractor follows an outward spiral close to the plane around an unstable fixed point. Once the trajectory spirals out enough, a second fixed point influences it, causing a rise and twist in the $z$ dimension. This behavior generates a quasi-periodic oscillatory pattern in the $x$ variable, with peaks and troughs in different amplitudes.
We apply the encoding approaches on two dynamical systems,  Lorenz  and Rossler systems (Figure \ref{fig:Lorenz_Rossler} (a), (d)).

\begin{figure*}[t]
	\centering
	\includegraphics[width=0.85\textwidth]{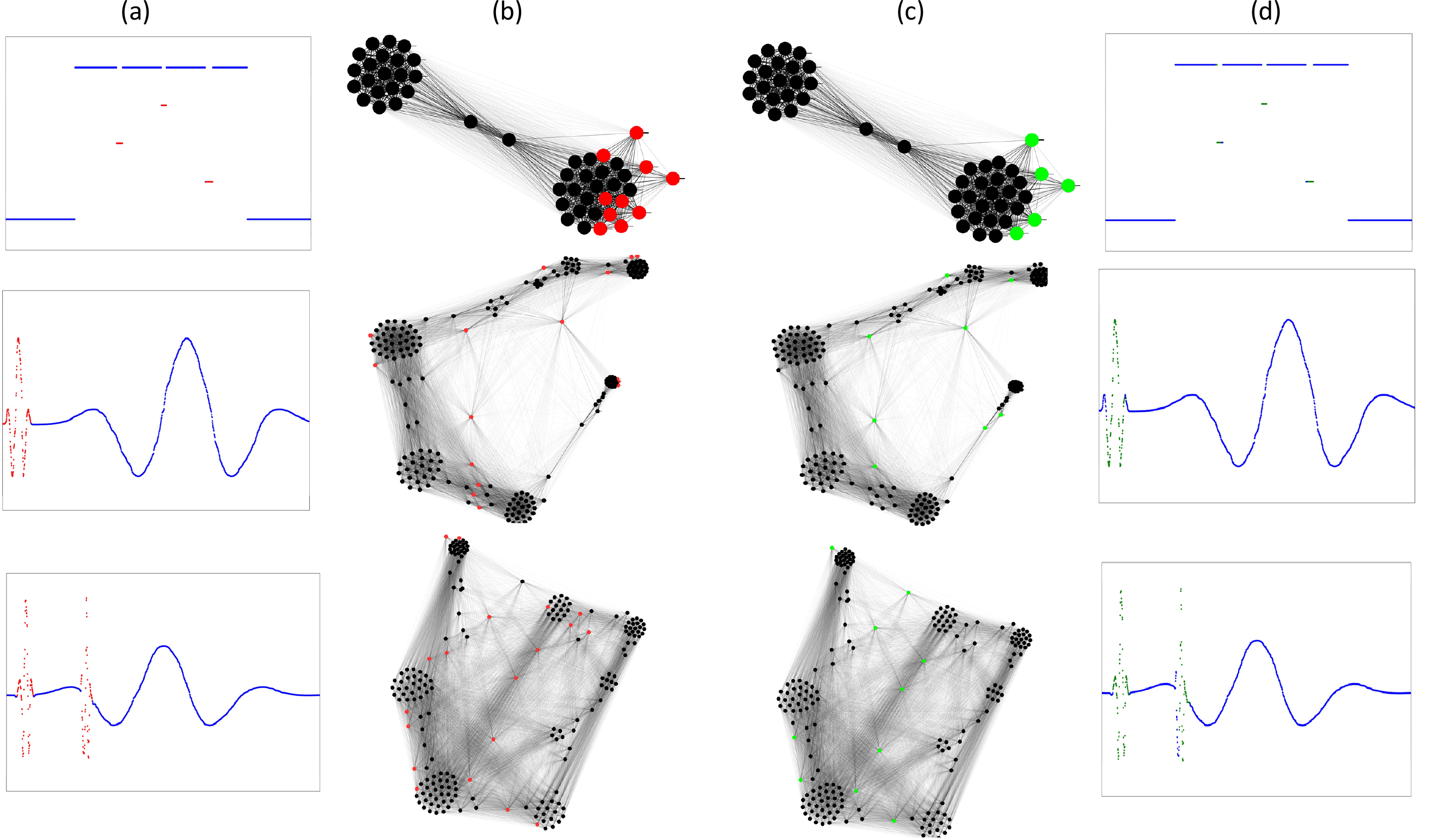}
	\caption{(a). Signals with the major pattern 1 (blue) and minor pattern 2 (red); (b). Generated graph, ground truth (outlier) is colored by red; (c). Vertex with pattern 1 (green) selected by visual inspection and the smallest clustering coefficients; (d). Inverse mapping from the graph to the original time series, the rare pattern in the time series subsequence obtained from the suspected vertex is colored by green.}
	\label{fig:anomaly_graph}
\end{figure*}

%From Figure \ref{fig:Lorenz_Rossler}, we can clearly identify the difference between these two dynamical systems. 
The network of Rossler system has a clear sparse connection area like a bipartite graph. The big blue vertices near the boarder of the bipartite gap indicate the homogeneous large transition towards the time end. This means the system has either converging or periodical behavior, which is consistent with the three even peaks at the ends, as the system begin to orbit in a quasi-periodical way. Lorenz system indicates strong coherence without any gap inside the graph. Instead, the big transition flow incurs from the start to the end as we can observe both enough big blue and red vertices. This means the system is very unstable, transiting among states very often. %without a converging or periodical trend. 

\begin{figure}[h]
	\centering
	\includegraphics[width=0.4\textwidth]{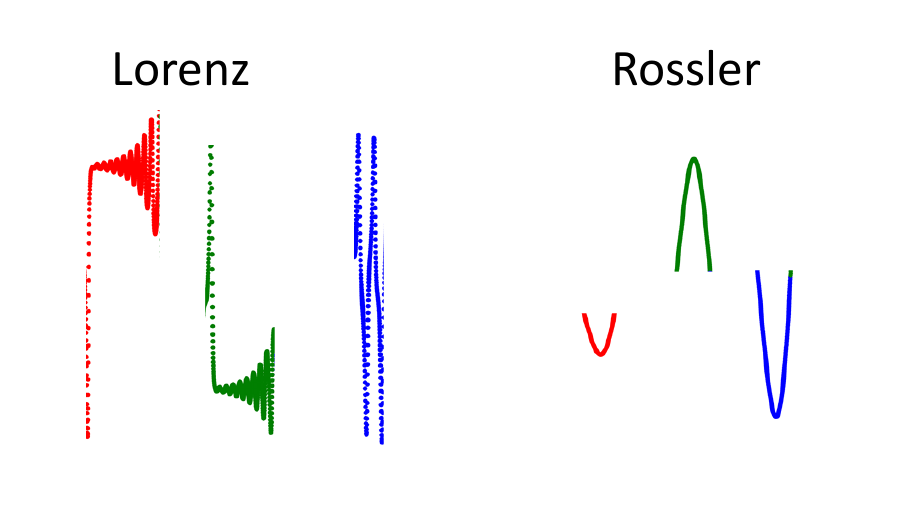}
	\caption{The basic time series "shapelets" discovered by the community in the graph.}
	\label{fig:shaplet}
\end{figure}

The most interesting results are the shapelet discovered from the graph communities, as shown in Figure \ref{fig:shaplet}. The community detection algorithm recognizes three modules in both systems. After matching the vertices with different module labels back to the raw time series, the shapelets share the common patterns respectively. For Lorenz system, three communities correspond to the high frequency oscillation (red), low frequency oscillation (green) and the alternating behavior switching between those two fixed points (blue).  Rossler system is quasi-periodic with steady slow transition among multiple magnitudes. The shapelet mapped from the community can be summarized as three basic 'wavelet' type: the convex wavelets within small (red) and big (blue) magnitude and the concave wavelet (green). Thus, the basic patterns embedded in the time series are discovered through the graph visualization and network statistics.    

\subsection{Anomaly Detection}
We test our approach on three typical compound functions \cite{wang2015adopting} with different rare patterns to test if we can identify those specific anomaly through the graph and network statistics. \footnote{More details about the generating function, please refer to \cite{wang2015adopting}.}

\begin{figure*}[t!]
	\centering
	\includegraphics[width=0.9\textwidth]{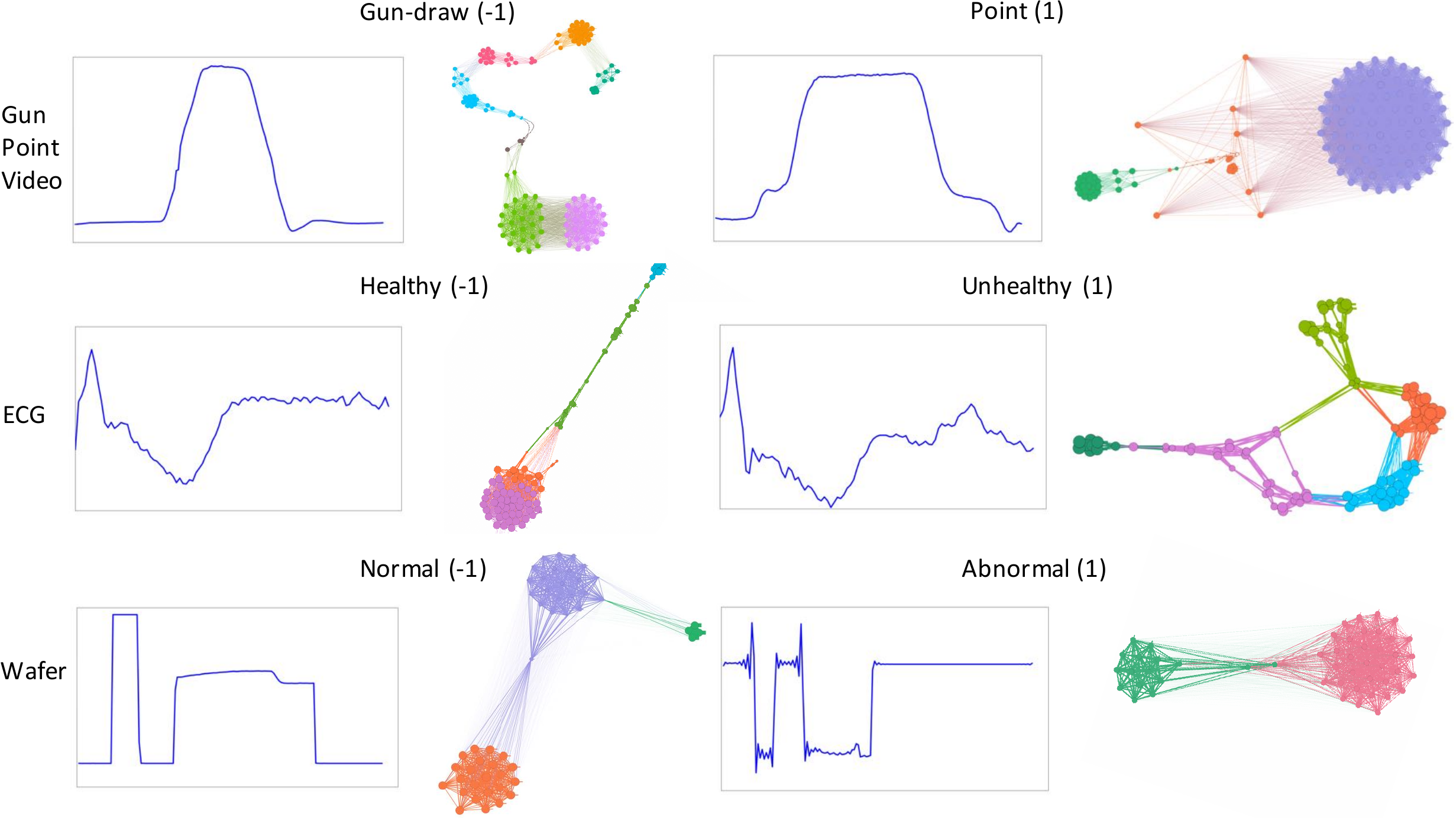}
	\caption{Original time series and the network graph with the modularity encoding.}
	\label{fig:clf}
\end{figure*}

%Briefly, generating functions represent the simple cases of three typical mixed modularity (Figure \ref{fig:anomaly_graph} (a)). The compound pattern in $F_1(x)$ are in the range of the uncontaminated data but appearing at wrong positions. This sometimes occurs when the value in a time series  is within normal magnitude but maybe corrupted or inaccurate. $F_2(x)$ simulates the mixed pattern with significant large deviations, indicating the situations where the time series contains some unknown samples with extraordinary bounds. $F_3(x)$ includes seemingly uncontaminated but shifted and under-sampled data. This may happen among heterogeneous observations from several experiments where the time series obtained is highly sparse and biased and the main function needs to get rid of their influence.

The network graphs (Figure \ref{fig:anomaly_graph}(b) and (c)) are generated with $Q=\{10, 10, 10\}$ and $S=\{50, 200, 200\}$ respectively. The ground truth of the primary pattern is highlighted in red. To separate the mixed structures generated by different functions in the graph, we select the vertices that are not bundled so closely within its near communities, since the different module has few but distinct patterns in the raw signal. The independent patterns isolate the corresponding vertices in the graph to be comparatively independent. The recognition process combines two procedures:

\begin{itemize}
	\item Select the most isolated vertices, denote as the set $\mathnormal{H}$.
	\item The top $k$ vertices with the smallest clustering coefficients are selected, denote as the set $\mathnormal{S}$
\end{itemize}

Then we use the union $\mathnormal{H} \cup \mathnormal{S}$ to finally decide which vertices are most isolated (green vertices in Figure \ref{fig:anomaly_graph}(c)). The vertices marked as from the different module are mapped back to the original time series to show the corresponding subsequence and their raw patterns. Comparing Figure \ref{fig:anomaly_graph} (a) and (d), our results demonstrate accurate recognition performance on where and what the rare patterns are.   

Note that sometimes the vertices selected by visual inspection and lowest clustering coefficient are significantly different. Choosing the set intersection $\mathnormal{H} \cap \mathnormal{S}$ will sometimes leads to the high false-negative rate as some important but rare patterns might be ignored. This would be the worst case in, for example,  anomaly detection. So we choose the union  $\mathnormal{H} \cup \mathnormal{S}$ to guarantee the low false-negative rate, as the false-positive patterns can be easily eliminated by inversely mapping the vertices to the original time series and comparing the subsequence distances or even by visually inspecting on those mapped shapelets.

\subsection{Classification}
Our final experiment is to identify the class labels on three real dataset: ECG from the monitoring sensors, Wafer recorded by the manufacturers and Gun-Point recorded by motion sensors \cite{wei2006semi}. The ECG dataset is a time series of the measurements recorded by one electrode during one heartbeat. The data has been annotated by cardiologists and a label of normal or abnormal is assigned to each data record. The Gun Point dataset contains two-dimensional time series extracted from motion sensor of two actors either aiming a gun or simply pointing at a target. The Wafer dataset is recorded by one vacuum-chamber sensor during the etch process of silicon wafers for semiconductor fabrication with the  normal or abnormal.

We randomly picked one sample of each label and draw the network graph with $Q = \{50, 50, 10\}$ for the Gun Point, ECG and Wafer dataset. Dimension reduction is skipped as we wish to keep full information in the original data. The network graph uses modularity encoding to highlight the distinct structures for different labels (Figure \ref{fig:clf}).

Even though we cannot observe small differences in the mechanical signals, network graphs can always show us the structural difference in their network layout. For ECG data, the healthy ECG has a linear structure with four close communities, while the unhealthy ECG tends to have a round structure with more communities. 

\begin{table}[h]
	\centering
	\caption{Classification accuracy of 1-NN classifier with the raw data and combined features.}
	\begin{tabular}{lrr}
		\toprule
		& \multicolumn{1}{l}{Acc (raw)} & \multicolumn{1}{l}{Acc (combined)} \\
		\midrule
		ECG   & 0.88  & 0.9 \\
		Wafer & 0.995 & 0.996 \\
		Gun-Point & 0.913 & 0.932 \\
		\bottomrule
	\end{tabular}%
	\label{tab:clfacc}%
\end{table}%\\

For the Gun-Point dataset, the signal is very similar as the Point pattern is stretched from the Gun-draw to have a wider plateau  in the middle. However, their network structures show a huge difference with seven and three clear communities, respectively.

\begin{table*}[t]
	\small
	\centering
	\caption{Net work statistics. On the ECG, Wafer and Gun Point dataset, the mean and standard deviation across all samples are reported. The bold number means difference of this measurement between labels are statistically significant in a paired T-test}
	\begin{tabular}{rrrrrrrrr}
		\toprule
		& Avg & AVG Weighted  & Diameter & Density & Modularity & \# Communities & AVG Clustering  & AVG Path  \\
		&  Degree & Degree & & & & &  Coefficient & length \\
		\midrule
		Lorenz & 62.06 & 1.00  & 2.00  & 0.99  & 0.18  & 3     & 0.96  & 1.05 \\
		Rossler & 48.97 & 1.01  & 2.00  & 0.78  & 0.41  & 3     & 0.87  & 1.25 \\
		3-Notch & 51.00 & 1.00  & 1.00  & 1.04  & 0.45  & 2     & 0.99  & 1.00 \\
		FineFeature & 130.22 & 1.01  & 2.00  & 0.65  & 0.71  & 7     & 0.83  & 1.36 \\
		UnderSample & 79.57 & 1.00  & 3.00  & 0.40  & 0.72  & 6     & 0.88  & 1.64 \\
		\midrule
		ECG   (-1) & \textbf{19.92} & \textbf{6.04} & 11.40 & \textbf{0.21} & 0.50  & \textbf{3.1} & 0.59  & 4.33 \\
		ECG   (1) & \textbf{13.42} & \textbf{4.97} & 10.00 & \textbf{0.14} & 0.58  & \textbf{5.2} & 0.43  & 3.93 \\
		Wafer (-1) & \textbf{41.36} & \textbf{1.03} & 2.00  & 0.84  & 0.47  & 2.2   & 0.94  & 1.20 \\
		Wafer (1) & \textbf{38.28} & \textbf{1.05} & 1.80  & 0.78  & 0.45  & 2.5   & 0.94  & 1.26 \\
		Gun Point (-1) & \textbf{33.85} & \textbf{19.79} & \textbf{19.80} & \textbf{0.23} & \textbf{0.64} & \textbf{6.8} & \textbf{0.80} & \textbf{8.48} \\
		Gun Point (1) & \textbf{56.07} & \textbf{36.16} & \textbf{13.20} & \textbf{0.38} & \textbf{0.44} & \textbf{2.9} & \textbf{0.85} & \textbf{6.22} \\
		\bottomrule
	\end{tabular}%
	\label{tab:stat}%
\end{table*}%

\subsubsection{Network Statistics and Summary}
%The encoding network graph not only supports the purpose of visualization of time series and senor signals, it also bridges the visual analytics, network dynamics and time series analysis with a natural inverse map from network graph to the original time series. 
Networks can be analyzed by exploring an extensive set of statistical properties of the associated time series with an active visual understanding. For example, motifs discovered in a network are mapped as specific spatial-temporal patterns in the raw signal, which are characterized by looking at the corresponding forward and backward maps. Different dynamical regimes in time series can also be visualized and understood by exploring an extensive set of topological statistics at the associated network domain.

Table \ref{tab:stat} summarizes the common network statistics on the experimental datasets \cite{bastian2009gephi}. On the real sensor signal like ECG, Wafer and Gun-Point, the average across all the samples with specific labels are reported. 

Specifically, some statistics associated with positive and negative labels are significantly different. The average degree and weighted degree are associated to distinguish between the negative and positive labels across all three dataset. This means that the network graph generated from each label has different overall network topology. The network density and the number of communities are also helpful in classifying healthy and unhealthy samples for the ECG signal. For the Gun-Point dataset, all statistics shows significant difference between the action of Gun-draw and Point. 

To quantify the quality of the network features, we combine the statistics in Table \ref{tab:stat} with the raw data and feed into the One-nearest-neighbor classifier. With the default training and test split, the classifier with the combined features outperform the one trained on the raw data (Table \ref{tab:clfacc}).

\section{Conclusion and Future Work}
In this work, we present a novel approach to translate time series into the network graph for visualization and pattern mining. %The proposed graph is able to encode both temporal ordering and statistical dynamics in the original time series through the Markov Transition Field (MTF). 
This encoding map has a natural inverse, enable the visual exploring of the interesting patterns in the generated graph and the inverse mapping to the original data. The experiments on both synthetic signals and real sensor data demonstrate the effectiveness of our approach. %The inverse mapping is one-to-one, which provides a link between pattern recognition in the network graph and time series analysis. Moreover, our work bridges the gap between visual analytics, time series data mining and graph statistics, as the the generated network can be analyzed by exploring an extensive set of statistical properties of the associated time series with an active visual understanding.

As one of the future work, we are interested in designing a unified user interface to build a set of interaction mechanism to enable the active exploring of the time series and sensor signal through the network graph by the users. This preliminary study is expected to enlighten more works on exploring the possibilities of using machine learning and visual analytics to create new solutions that exploit the sensor data, improve the human computer interaction in a more transparent and controllable IoT framework.

\newpage
\bibliographystyle{aaai}
%%use following if all content of bibtex file should be shown
%\nocite{*}
\bibliography{template}

\end{document}